\documentclass{article}

\usepackage{PRIMEarxiv}
\usepackage[section]{placeins}
\let\Oldsubsection\subsection
\renewcommand{\subsection}{\FloatBarrier\Oldsubsection}

\usepackage[utf8]{inputenc} 
\usepackage[T1]{fontenc}    
\usepackage{newunicodechar}
\newunicodechar{́}{\'}
\usepackage{hyperref}       
\usepackage{url}            
\usepackage{booktabs}       
\usepackage{amsfonts}       
\usepackage{nicefrac}       
\usepackage{microtype}      
\usepackage{lipsum}
\usepackage{fancyhdr}       
\usepackage{graphicx}       
\usepackage{amsmath}
\usepackage{multirow}
\graphicspath{{media/}}     

\title{\textbf{A Prototypical Network with an Attention-based Encoder for Drivers Identification Application}}

\author{
  Wei-Hsun Lee$^{1,*}$,~\textit{Senior Member, IEEE},
  Kuang-Yu Li$^{2}$, 
  and Che-Yu Chang$^{1}$ \\[3pt]
  $^{1}$Dept. of Transportation \& Communication Management Science, 
  National Cheng Kung University, Taiwan\\
  $^{2}$Institute of Data Science, 
  National Cheng Kung University, Taiwan\\[3pt]
  *Corresponding author: Wei-Hsun Lee (e-mail: \texttt{leews@mail.ncku.edu.tw})
}

\begin{document}
\maketitle

\begin{abstract}
Driver identiﬁcation has become an area of increasing interest in recent years, especially for data-driven applications, because biometric-based technologies may incur privacy issues. This study proposes a deep learning neural network architecture, an attention-based encoder (AttEnc), which uses an attention mechanism for driver identiﬁcation and uses fewer model parameters than current methods. Most studies do not address the issue of data shortages for driver identiﬁcation, and most of them are inﬂexible when encountering unknown drivers. In this study, an architecture that combines a prototypical network and an attention-based encoder (P-AttEnc) is proposed. It applies few-shot learning to overcome the data shortage issues and to enhance model generalizations. The experiments showed that the attention-based encoder can identify drivers with accuracies of 99.3\%, 99.0\% and 99.9\% in three different datasets and has a prediction time that is 44\% to 79\% faster because it signiﬁcantly reduces, on average, 87.6\% of the model parameters. P-AttEnc identiﬁes drivers based on few shot data, extracts driver ﬁngerprints to address the issue of data shortages, and is able to classify unknown drivers. The ﬁrst experiment showed that P-AttEnc can identify drivers with an accuracy of 69.8\% in the one-shot scenario. The second experiment showed that P-AttEnc, in the 1-shot scenario, can classify unknown drivers with an average accuracy of 65.7\%.  
\end{abstract}

\section{Introduction}
Modern vehicles are equipped with many sensors and controllers, such as OBD-II, GPS, and advanced driver assistance system (ADAS) equipment, to retrieve information about vehicle dynamics. The rapid development of the Internet of Vehicles (IoV) and in-vehicle systems brings big data to the transportation industry in order that big data analysis and deep learning techniques can be used to develop applications such as detecting driving fatigue, analyzing driving styles and identifying drivers.

\setlength{\parindent}{2em}Most current driver identification technologies use biometrics, such as face recognition, iris recognition, and fingerprint recognition. Some of these have advantages; for example, face recognition can be used for driver identification and can also be used to detect fatigue in drivers. Data-driven driver identification is not yet popular, but most biometric-based driver identification systems have issues in terms of cost, privacy, and driver friendliness, which are addressed by data-driven driver identification. Data-driven driver identification continuously classifies time series vehicle dynamic data to learn the driving characteristics of a driver, and it is not a one-off identification procedure.

\setlength{\parindent}{2em}Data-driven driver identification is a time series classification technology that can be applied to several areas in transportation. For example, driver identification can be used to prevent theft and enable driver-personalized services, and car-sharing services. The driver identification technique can be applied to assign responsibilities for an accident or damage, and it can be used for fleet management to determine whether a driver violates shift rules. It can also be used to prevent car owners from registering a car in the name of a family member to avoid insurance costs.

\setlength{\parindent}{2em}Recurrent neural networks (RNNs) have been widely used to learn patterns from time series for decades, especially for driver identification. However, some studies show that when RNN is used in long sequences, performance is limited because it uses the output from the previous step and cannot capture long-term relationships. Long short-term memory (LSTM) and gated recurrent units (GRUs) address these shortcomings of RNNs. They perform better on long-term sequences, but the need for large parameters increases the risk of overfitting. This study uses an attention mechanism to construct an attention-based encoder to reduce the number of model parameters that are required to identify drivers.

\setlength{\parindent}{2em}Data shortages are a critical issue for driver identification due to the unwillingness of drivers to participate in data collection. More seriously, only a small portion of the collected data can reflect the distinctiveness of the driver, e.g., idling and cruising. Such problems amplify the data shortage issue in driver identification. Most related works do not address these problems, which renders these methods infeasible in the real world. In this paper, few-shot learning is used to address the problem of data shortages, which is widely used for image recognition ([1], [2], [3]) but rarely for time series.

\setlength{\parindent}{2em}We think that a driver identification model can be applied as follows:

a. Anti-theft System: Detect the intruders of a family car.

b. User-Based Insurance: Detect whether the rental car is driven by one or more people to adjust the insurance amount.

c. Fleet Management: Check whether the driver works in regular shifts.

d. Personalized Service: Adjust seats or play specific music according to drivers’ preferences.

\setlength{\parindent}{2em}For the applications listed above, some of them cannot be addressed by current driver identification methods, which limits the various applications of driver identification. For example, an anti-theft system needs to detect drivers that are unknown, but most of the current methods can classify "known" drivers only. For the application of usage-based insurance (UBI), supposing a rental car is driven by thousands of different driver groups, costs are increased if the driver identification method cannot classify unknown drivers. In the following sections, we pinpoint the disadvantages of current solutions to driver identification and then propose remedies. 

\subsection{Problem Description}
1) Data shortage
Machine learning methods have used big data successfully, but they may collapse when data shortages occur, which is a real problem in the real world. The unwillingness of drivers to participate in data collection and the small amounts of collected data available to reflect the distinctiveness of the driver are the reasons for data shortages. Although the current driver identification methods can achieve high accuracy, most of them do not discuss the problem of data shortages.

2) Number of Model Parameters
RNN, LSTM and GRU are most commonly used to learn time series and to classify drivers [4], [5], [6], [7]. These studies classify drivers with high accuracy, but the large number of model parameters increases the cost of data collection and the required computer capacity to result in good prediction capabilities.

3) Applying to Various Data Sources

Most studies use variable selection methods and feature engineering approaches to achieve greater accuracy ([8], [9], [10], [11], [12]). These studies require expert knowledge, and the computational costs increase with the number of model parameters. Most importantly, whether their methods can be applied to different data sources is still unknown.

4) Model Generalization
Driver identification methods should be flexible in various application scenarios. However, most methods are mainly based on traditional classification algorithms, so they may be able to classify unknown drivers. To address this problem, we usually use retraining or transfer-learning methods but these two methods are expensive. To date, flexible driver identification methods are still lacking.

\subsection{Motivation}
This study improves existing driver identification methods. In terms of data preprocessing, this study reduces the cost of feature engineering. RNN is commonly used for time series learning; however, it incurs a growth in the number of model parameters. In this study, we develop a more general neural network structure that uses an attention mechanism to reduce the parameters needed for driver identification. Furthermore, few-shot learning is considered to overcome the issue of data shortages and make the model more generalized.

\setlength{\parindent}{2em}The main contributions of this work are:

a. The cost of feature engineering and the number of variables for vehicle dynamic data are largely reduced to allow more practical driver identification applications.

b. Drivers are identified using the proposed attention-based encoder (AttEnc, as shown in Fig. 2), which uses an attention mechanism that requires fewer model parameters.

c. A more general model, an attention-based encoder with a prototypical network (P-AttEnc, as shown in Fig. 3), is proposed to address the issue of data shortages and the ability to classify unknown drivers.

The remainders of the paper are organized as follows: Section II reviews related works for driver identification and few-shot learning technologies. Section III presents the methodology and discusses the details of the proposed model, AttEnc and P-AttEnc. The datasets and the preprocessing procedure of the experiments are presented in Section IV, and competitors of our model and training settings are detailed. The results for AttEnc and P-AttEnc are shown in Section V. We explain the good performance of our model and visualize the driver fingerprints that are extracted from P-AttEnc using t-SNE techniques to verify our models. Section VI draws conclusions and details future work.

\section{Related Works}
This section first details related studies for methods of driver identification and then shows how few-shot learning can be used for data shortage issues to improve driver identification technologies.

W. Dong \textit{et al. }[7] used GRU to extract features from GPS data and combined supervised learning and unsupervised learning as ARNet. Fifty-driver classification experiments gave an accuracy of 40.4\% on segmentation and 78.3\% had top-5 accuracy. A. Girma \textit{et al. }[4] used an end-to-end LSTM\_RNN architecture as a driver identification model. This reduces the number of stages of feature engineering because of the ability of LSTM to learn time series. This study mainly focused on noise expeiments. For increased noise and outliers, the method is more than 88\% accurate. J. Chen \textit{et al. }[12] used an unsupervised three-layer nonnegativity-constrained autoencoder to search for the best window size and constructed a deep nonnegativity-constrained autoencoder for driver identification. This method achieves 99.65\% accuracy using a thirty second time window, but it requires a deeper network so that the number of parameters can increase significantly, and it is difficult to determine the window size.

J. Zhang \textit{et al. }[5] proposed a model that uses a convolutional neural network (CNN), RNN, and attention mechanism for driver classification. This model learns time series without the need for additional feature engineering steps and better learns relationships between multidimensional features. The experiment used the Ocslab driving dataset of ten drivers and achieved the best accuracy of 98.36\% for a window size of sixty seconds and a six-second overlap. CNN+RNN architecture has been used to learn time series in other fields, e.g., [13], [14], and [15]. The attention mechanism has been proven to better learn sequences, and some studies [16], [17], [18] also use it with a CNN+RNN architecture to increase effectiveness.

Few-Shot Learning is rarely used for time series. Currently, it is mostly used to address the issue of data shortages in the imaging field. N. Goel \textit{et al. }[2] and G. Song \textit{et al. }[1] noted the difficulty of real-world data collection and used a prototypical network for font recognition and food recognition. Some driver identification studies use few-shot learning. H. Dang \textit{et al. }[19] used a Siamese network for driver verification. Their study considered driver verification as a comparison of embedding similarity, and obtained an average AUC of 0.753 in the experiments. S. H. Sánchez \textit{et al. }[20] converted twenty-five drivers’ acceleration data into 2D inputs and used ResNet50+GRU to classify drivers with a top-1 accuracy of 71.89\%. The study also used a Siamese network for driver verification, similar to the study of H. Dang \textit{et al. }[19], and achieved a 74.09\% f1 score. Although these two studies are similar to our work, they solve a verification problem using a Siamese network, so they actually predict whether the car owner can drive a car, which is different from our study. These two studies used few-shot learning, which is rare for driver identification; moreover, they did not address the issue of data shortages.

In summary, most related works use neural network architecture to learn time series, which allows accurate driver identification. Neural network architecture is less dependent on feature engineering than machine learning methods but most studies still use RNN to learn, so the number of model parameters is still a significant problem. Finally, few studies address the issue of data shortages for driver identification. 

\section{Methodology}
This study proposes a two-stage process for driver identification. Stage 1 uses a neural network structure called an attention-based encoder (\textit{AttEnc}) to identify drivers using fewer model parameters than CNN- or RNN-based models. Stage 2 uses an attention-based encoder with a prototypical network called \textit{P-AttEnc }to generalize \textit{AttEnc}. It has the ability to classify both "known" and "unknown" drivers and is able to train in few-shot scenarios.

\begin{figure}
    \centering
    \includegraphics[width=0.5\linewidth]{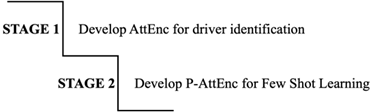}
    \caption{Two-stage classification is used to solve the issues of driver classification and data shortages. }
    \label{fig:placeholder}
\end{figure}

\subsection{Attention-based Encoder (ATTENC)}
The idea of an attention-based encoder was proposed by Song et al. [21], but the proposed model in this study is different in two respects. The first is positional embedding, [21] calculates sine and cosine functions as positional information for each time point. This study attempts to reduce the number of parameters, so a simple embedding layer is used to provide positional information for \textit{AttEnc}. Second, the previous study [21] used dense interpolation to avoid dimensional explosion and to better capture the structure of natural language sentences. This study uses a fully connected layer to achieve good training results with an easier implementation. 

\begin{figure}[htbp]
    \centering
    \includegraphics[width=0.5\linewidth]{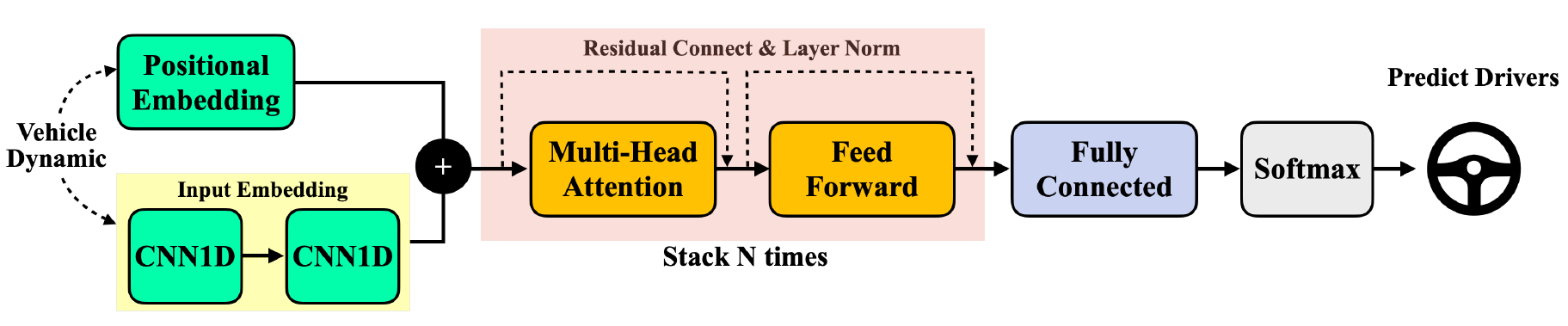}
    \caption{Attention-based Encoder for driver identification.}
    \label{fig:placeholder}
\end{figure}

The whole AttEnc process involves preprocessing raw data and normalizing them before feeding data into the model first. Then, to add an attention mechanism, our study uses two layers of one-dimensional convolution (CNN1D) to project the multidimensional time series input to the feature space. Attention does not have the concept of position and location. The attention mechanism can focus the weight of the network on the meaningful pixels, promote the expression of effective features and capture the relationships on the sequence with only one step. It can improve the efficiency of training. Feed forward means that the output of each layer of calculation will only be forwarded to the input of the next layer in a single direction; that is, the input and output are independent. Therefore, the meaningful features of driver identification will be captured after stacking n times. Finally, a fully connected layer and softmax layer are used to identify driving.

The neural network structure of the proposed attention-based encoder (\textit{AttEnc}) model is illustrated in Fig. 2, which is composed of five parts. The physical meanings of each part in the attention-based encoder are briefly described as follows:

1) Positional Embedding
At each time point of a vehicle dynamic window, vectors are used to express the relationship within the series.

2) Input Embedding
A 1D convolution network (CNN1D) is used to extract meaningful features and to express relationships between dimensions in a vehicle dynamic window. (The generated features are used only latently and internally in the model). Two layers of 1D convolution are stacked in our model.

3) Multi-Head Attention
The attention mechanism applied the method of A. Vaswani et al. [22] to define the interrelationships in the sequence by only one step, as Eq. (1). 

\begin{equation}
Attention(Q, K, V) = Softmax\!\left(\frac{QK^{T}}{\sqrt{d_K}}\right)V
\end{equation}
More sequence information is obtained using different dimensions of attention heads, as Eq. (2):

\begin{equation}
\begin{aligned}
\text{MultiHead}(Q, K, V) &= \text{Concat}(h_1, \ldots, h_h) W^O, \\
\text{where} \quad 
h_i &= \text{Attention}(Q W_i^Q,\, K W_i^K,\, V W_i^V)
\end{aligned}
\label{eq:multihead}
\end{equation}

The attention mechanism can make the network focus on the prominent areas of the feature map. Therefore, all the collected data are fed into the AttEnc model so that there is no explicit variable to represent the features in the algorithm.

4) Residual Connect
This increases the sensitivity of neural networks to small changes and is widely used for deep neural networks.
5) Layer Normalization
Data are normalized along the feature dimension to allow more stable training for small batch data and sequential networks. 

\subsection{Prototypical Network with Attention-based Encoder (P-ATTENC)}
In Stage 2, \textit{P-AttEnc }(Fig. 3) uses the N-way K-shot prediction method from few shot learning, and the training data and testing data are composed of a support set and query set.

\begin{figure}[htbp]
    \centering
    \includegraphics[width=0.5\linewidth]{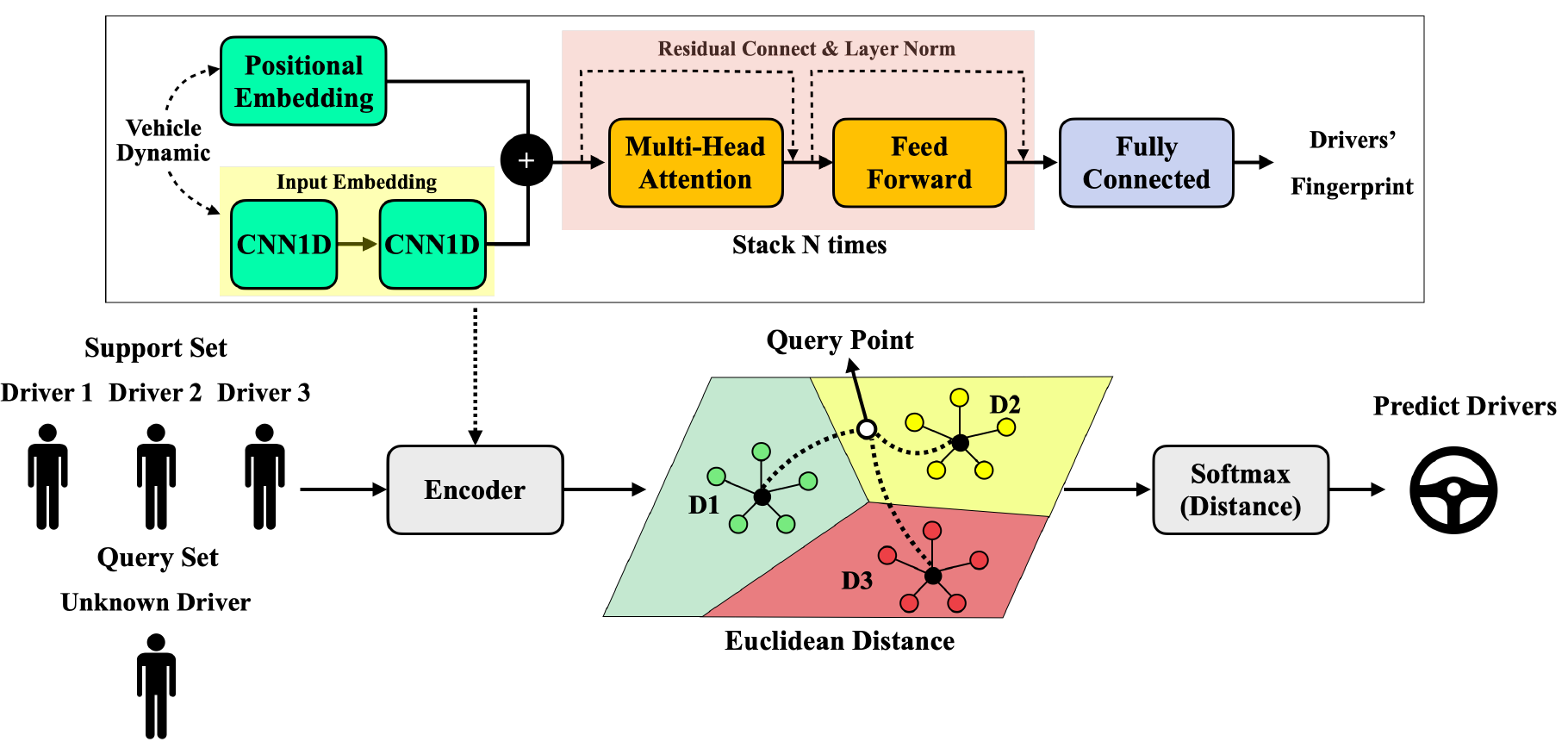}
    \caption{Prototypical Network with Attention-based Encoder for Few-Shot Learning. }
    \label{fig:placeholder}
\end{figure}

To be clear, an N-way K-shot scenario represents an episode, and the support set and the test set both contain various N-way K-shot episodes, where N-way represents the number of categories, and K-shot represents the number of data in each category. In the training phase of \textit{P-AttEnc}, we use the \textit{AttEnc} $f_{\phi}$ (Eq.~3).

\begin{equation}
F_{\phi} : \mathbb{R}^D \rightarrow \mathbb{R}^M
\end{equation}
To extract embeddings of each N-way K-shot episode from the support set of training data to M dimensions. In each episode, we calculate the mean embeddings for each driver as their prototype (Eq. 4) and use the query set from training data to update \textit{P-AttEnc}.

\begin{equation}
C_K = \frac{1}{|S_K|} \sum_{(x_i, y_i) \in S_K} F_{\varphi}(x_i)
\label{eq:prototype}
\end{equation}
Later, in the testing phase of P-AttEnc, AttEnc is still used to extract embeddings of each N-way K-shot episode from the support set of testing data and calculate the mean embeddings for each driver to obtain the prototype. After that, the Euclidean distance is used to calculate the distance from the query points \textit{X} to each prototype, and the nearest one is considered the result. Softmax function (Eq. 5) is used to normalize the distances to obtain probabilities.
\begin{equation}
p_{\varphi}(y = K \mid X) =
\frac{\exp\!\left(-d\big(f_{\varphi}(X), C_K\big)\right)}
{\sum_{K'} \exp\!\left(-d\big(f_{\varphi}(X), C_{K'}\big)\right)}
\label{eq:proto_prob}
\end{equation}
Finally, P-AttEnc is optimized by minimizing the negative log-probability (Eq. 6).
\begin{equation}
J(\varphi) = -\log p_{\varphi}(y = K \mid X)
\label{eq:loss}
\end{equation}

\section{Experiments}
\subsection{Data Description}
The experiments in this study apply three open datasets: the OcsLab driving dataset [23], the Vehicular data trace [24] and the hciLab driving dataset [25], where each of the datasets has ten drivers. The data sources include the raw data collected from CAN-bus, GPS, and inertial measurement unit (IMU) sensors. These datasets are natural driving study datasets (NDS), so no human-controlled factors affect the experiment, except for the prescribed driving route. Detailed information about these three datasets is shown in Table 1.

\begin{table}[htbp]
\centering
\caption{Detailed information about the experimental datasets}
\label{tab:dataset_info}
\begin{tabular}{lccc}
\hline
\textbf{Info.} & \textbf{OcsLab} & \textbf{Vehicular} & \textbf{hciLab} \\ 
\hline
Number of drivers & 10 & 10 & 10 \\ 
Data Source       & CAN-bus & CAN-bus & GPS + xyz Acceleration \\ 
Driving Time      & 23 hrs & 28 hrs & 0.5 hrs \\ 
Driving Length    & 46 km & 36 journeys & 23.6 km \\ 
\hline
\end{tabular}
\end{table}
\subsection{Data Preprocessing}

Overall computational time and prediction accuracy are the two criteria to evaluate the effectiveness of driver identification. The computing time depends on the data preprocessing workflow, and the prediction accuracy depends on the distinctiveness of the data features and model architecture. To ensure that the proposed method is effective for most data sources, different preprocessing methods are used for different data sources. There is no preprocessing if the data source is CAN-bus because it is assumed that CAN-bus data have sufficient features for the model to learn. If the data source is GPS with three-axis acceleration, basic statistical features (minimum, maximum, mean, 25th quantile, 50th quantile, and 75th quantile) are calculated. Many studies [8], [26], and [27] use these features to extract distinctive driving behavior. 

Three vehicle dynamic data sources are applied in this paper: the CAN-bus, IMU, and GPS. Vehicle dynamic data are a kind of multidimensional time series, and each dimension represents data collected from different sensors of the on-board telematics system, such as speed, acceleration, bearing, yaw, roll, and pitch. Normalizing data is a common step before feeding data into deep learning models, as it prevents each feature from being at different scale levels. In this study, the MinMax normalization method is applied to scale the original data into the range [0, 1] without changing its original distribution, making it easier for the model to learn uniform patterns. The normalization process is expressed in Eq. (7): 

\begin{equation}
x'_i = \frac{x_i - \min(x)}{\max(x) - \min(x)}
\label{eq:minmax}
\end{equation}
To preserve contextual information and ensure data continuity, the preprocessed data are further divided into overlapping time windows. Following the method of W. Dong et al. [27], a 30-second window with 50\% overlap is used. After completing data preprocessing, normalization, and window slicing, the processed inputs for the AttEnc and P-AttEnc models are obtained.

In Stage 1, the proposed AttEnc model is used as a classifier and compared with other common feature extractors. An attention mechanism combined with CNN and RNN is applied as the baseline for comparison, as it has demonstrated effectiveness in other domains. Additionally, the proposed method is compared to the autoencoder-regularized network (ARNet) proposed by [7], and the number of model parameters across models is examined to assess prediction performance.

\subsection{Experiment Setup}
The Stage 1 experiment is designed to evaluate the accuracy and efficiency of driver identification of the proposed \textit{AttEnc }model. For the purpose of comparison, a CNN+RNN with an attention mechanism architecture (i.e., CoLSTM\_Att and CoGRU\_Att) and an autoencoder regularized network (ARNet) are chosen as the three compared models in the experiment. Three open datasets are used, and each dataset has ten drivers: OcsLab, hciLab and Vehicular. Eighty percent of the data were used for training and twenty percent for testing, and each model was subjected to 5-fold cross validation to acquire the accuracy and the standard deviation. The experimental results are presented in Table 2, where the proposed \textit{AttEnc }outperforms the other three compared models, CoLSTM\_Att, CoGRU\_Att and ARNet, in accuracy and efficiency in all three datasets.

\textit{AttEnc }uses a layer of sixteen-head MultiHead attention and is stacked together one time with the feedforward layer. For the CNN+RNN with an attention mechanism architecture, we only stack a layer of CNN and a layer of LSTM or GRU to reduce the number of model parameters. Each model uses an Adam optimizer with a learning rate of 0.001 and is trained for 150 epochs with a batch size of 32, except for ARNet, which uses the same settings as in [7].

We design two Stage 2 experiments to overcome the data shortage issue and to enhance model generalization, that is, to evaluate the performance of the proposed \textit{P-AttEnc }model under few shot scenarios. In this stage, \textit{AttEnc }is used as an encoder for the prototypical network, and N-way K-shot prediction is adopted. For all the experiments, the training data are eighty percent of the datasets. We construct episode learning by using training data, and an episode represents an N-way K-shot scenario. Each epoch consists of two hundred episodes, and \textit{P-AttEnc }is trained for fifty epochs.

To show the flexibility of \textit{P-AttEnc}, two experiments are included in this stage. The first experiment trains \textit{P-AttEnc }in 10-way scenarios and tests it in 5-way and 10-way scenarios, and the drivers that appear in testing episodes are included in the training data. The OcsLab driving dataset and hciLab driving dataset are used in this experiment. The second experiment trains \textit{P-AttEnc }in 6-way, 7-way and 8-way scenarios and tests it in 2-way scenarios. We randomly choose N drivers from the training data to construct training episodes and randomly choose two drivers in the 10-N drivers to construct testing episodes. The drivers that appear in the testing episodes are not included in the training data; that is, the drivers in the testing data are considered unknown drivers. Only the OcsLab driving dataset is used in this experiment.

\section{Evaluations}
\subsection{Result of Experiment (Stage 1)}
We compare it with the related work discussed in Section II, including LSTM (A. Girma \textit{et al. }[4]), GRU (J. Zhang \textit{et al. }[5]) and ARNET (W. Dong \textit{et al.}[7]), which are proposed by other authors. To be fair, we add an attention mechanism to the models and compare them with the model we proposed to ensure that the final results are the most credible.
Driver identification is evaluated in terms of accuracy and standard deviation for 5-fold cross validation. The model parameters for each model and CPU time performance for the testing set are shown in Table 2.

\begin{table}[htbp]
\centering
\caption{Results for Driver Identification using Each Dataset.}
\label{tab:results}
\begin{tabular}{l l l l l}
\hline
\textbf{Dataset} & \textbf{Model} & \textbf{Accuracy} & \textbf{Params.} & \textbf{CPU time (s)} \\
\hline
\multirow{4}{*}{OcsLab} 
& AttEnc & 99.3(0.39) & 31,162 & 0.11 \\
& CoLSTM\_Att & 96.6(0.54) & 213,178 & 0.20 \\
& CoGRU\_Att & 97.4(0.44) & 163,770 & 0.25 \\
& ARNet & 89.3(1.84) & 1,484,604 & 0.40 \\
\hline
\multirow{4}{*}{hciLab}
& AttEnc & 99.0(0.03) & 27,066 & 2.18 \\
& CoLSTM\_Att & 99.0(0.01) & 209,082 & 6.64 \\
& CoGRU\_Att & 99.0(0.02) & 159,674 & 6.32 \\
& ARNet & 99.0(0.02) & 1,472,316 & 10.29 \\
\hline
\multirow{4}{*}{Vehicular}
& AttEnc & 99.9(0.10) & 37,338 & 0.10 \\
& CoLSTM\_Att & 99.5(0.36) & 208,570 & 0.18 \\
& CoGRU\_Att & 99.5(0.14) & 159,162 & 0.18 \\
& ARNet & 89.6(1.00) & 1,484,604 & 0.37 \\
\hline
\end{tabular}
\end{table}
The experimental results show that \textit{AttEnc }provides very accurate driver identification. \textit{AttEnc }is the most accurate method for the OcsLab dataset and the Vehicualr dataset (99.3\% and 99.9\%) because MultiHead attention better learns time series. CoLSTM\_Att and CoGRU\_Att perform well, but LSTM and GRU use the output of the previous step, which results in five times more parameters than that used in \textit{AttEnc}. For model parameter comparisons, each model was tested using twenty percent of the dataset, and the CPU calculation time was recorded. \textit{AttEnc }has the fastest CPU computing time because \textit{AttEnc }uses the fewest model parameters; in contrast, ARNet has the slowest CPU calculation time and the lowest accuracy. The 3D histogram comparing the number of model parameters is shown in Fig. 4, and the CPU time comparison is shown in Fig. 5.

\begin{figure}[!hbt]
    \centering
    \includegraphics[width=0.5\linewidth]{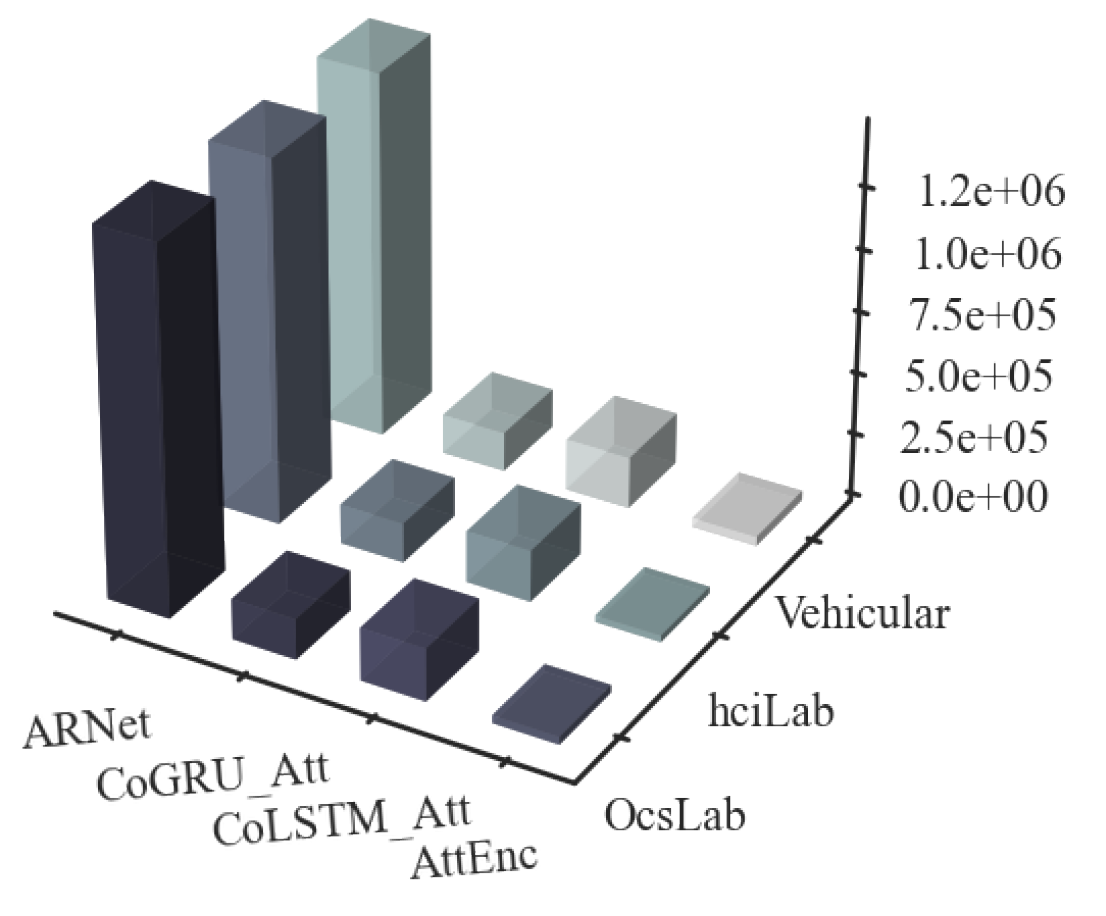}
    \caption{Performance Evaluation in Four Driver Identification Models: Parameters Comparisons}
    \label{fig:placeholder}
\end{figure}

\begin{figure}[!hbt]
    \centering
    \includegraphics[width=0.5\linewidth]{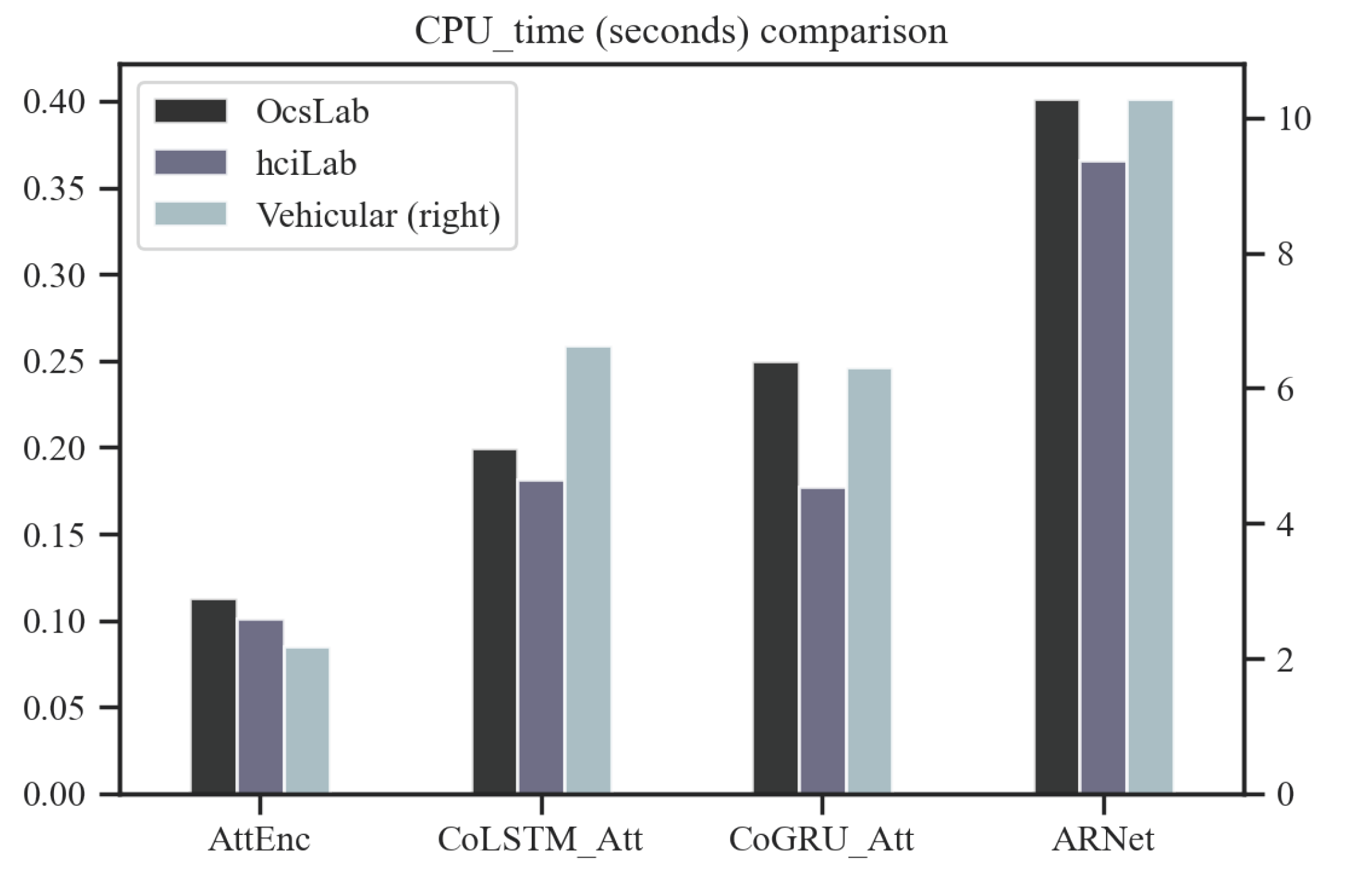}
    \caption{Performance Evaluation in Four Driver Identification Models: CPU time}
    \label{fig:placeholder}
\end{figure}

\subsection{Result of Experiment (Stage 2)}

\textit{P-AttEnc }uses \textit{AttEnc }to extract features from the vehicle dynamic data, and a prototypical network is then applied to classify drivers using the N-way K-shot method. To determine the generalization ability of \textit{P-AttEnc}, different settings of N-way K-shot scenarios were performed to simulate different conditions in the real world.
\begin{figure}[htbp]
    \centering
    \includegraphics[width=0.5\linewidth]{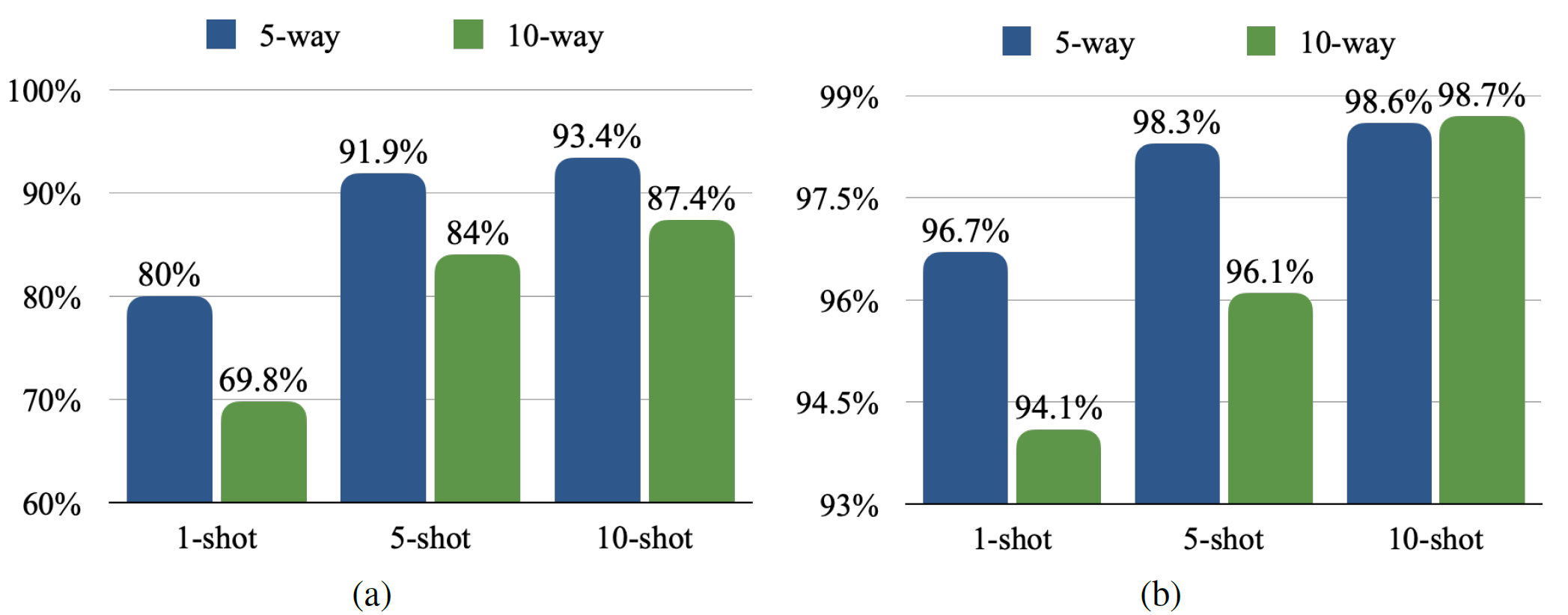}
    \caption{Results of experiment I in stage 2 on two datasets, (a)OcsLab dataset and (b)hciLab dataset.}
    \label{fig:placeholder}
\end{figure}

Fig. 6(a) and Fig. 6(b) show the results on the two datasets on which P-AttEnc is trained in 10-way and tested in 5-way and 10-way scenarios. It shows that if the “shot” is higher, \textit{P-AttEnc }gives a more accurate prototype, which results in a higher accuracy. If the “way” is greater, \textit{P-AttEnc }is slightly less accurate, but the results show that \textit{P-AttEnc }is very good in terms of generalization because it is trained using limited training data. We obtained the highest accuracy of 93.4\% using the OcsLab dataset for a 5-way 10-shot scenario and 98.7\% accuracy using the hciLab dataset for a 10-way 10-shot scenario. These results are comparable to those using sufficient data.

A visualization analysis is included in the first Stage 2 experiment. We extracted an episode from a 10-way 10-shot experiment and averaged the driver fingerprints (embeddings) to obtain the prototype (i.e., mean of the K-shot) for each driver. t-SNE ([28]) is used to project the driver fingerprints that are extracted by P-AttEnc into 2D space for visualization purposes. As shown in Fig. 7, the driver fingerprints for different drivers are relatively separated from each other, and the fingerprints for the same driver are near their own pro-
totype (marked with ’X’). Moreover, driver fingerprints for the hciLab driving dataset (Fig. 7(b)) are more distinctive than those for the OcsLab driving dataset (Fig. 7(a)).

\begin{figure}[htbp]
    \centering
    \includegraphics[width=0.5\linewidth]{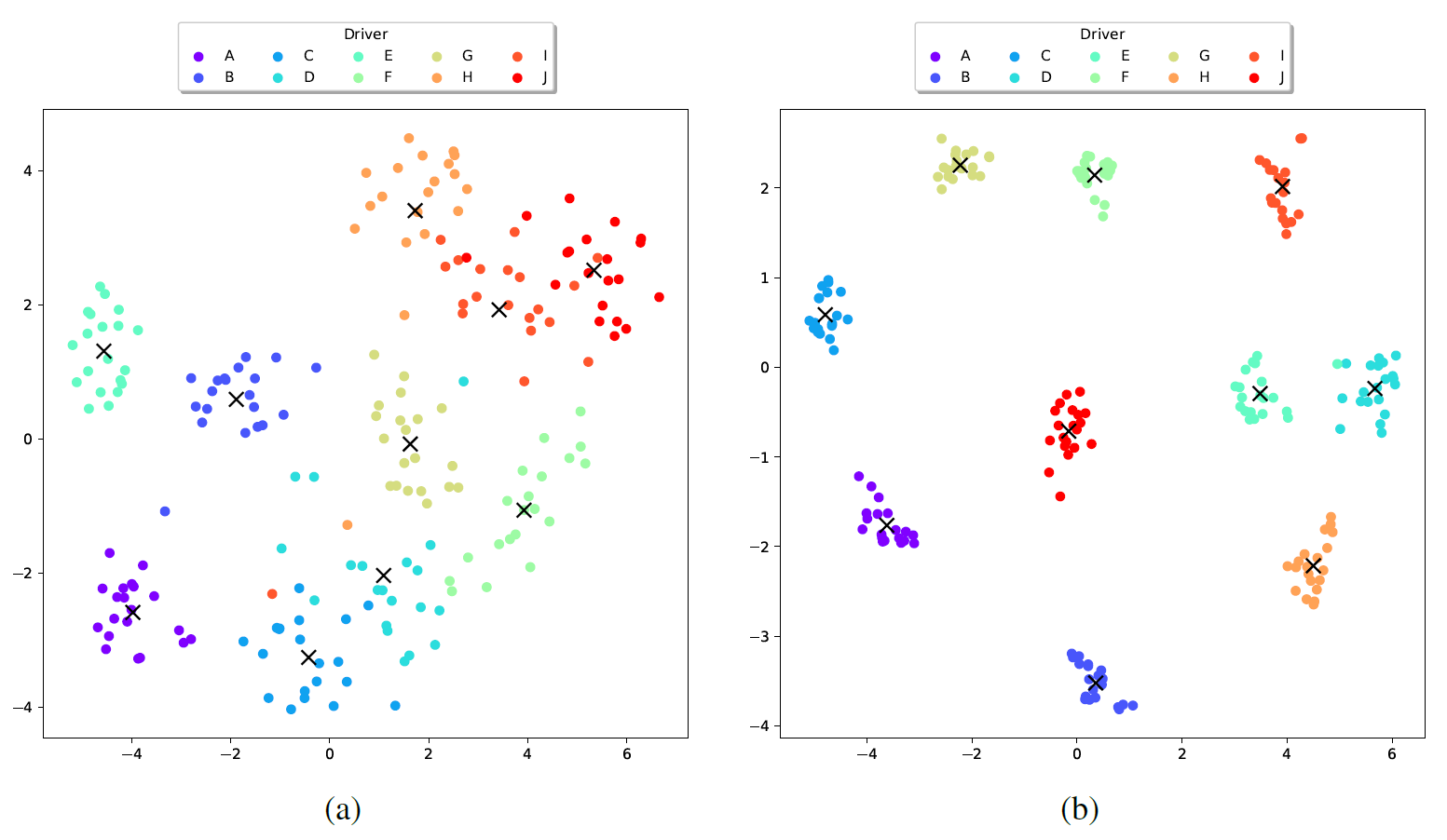}
    \caption{A sample of t-SNE visualization of a 10-way 10-shot episode and driver prototype (cross) on two datasets, (a)
is from OcsLab dataset and (b) is from hciLab dataset.}
    \label{fig:placeholder}
\end{figure}

This result, inferred that the hciLab driving dataset, created more representative features by calculating statistical features during the data preprocessing steps. Under extreme data shortage circumstances (i.e., 10-way 1-shot scenario), the experimental results show that P-AttEnc can be used to extract distinctive features and to classify drivers with an accuracy of more than 69.8\% for the OcsLab driving dataset.
 
Fig. 8 shows the results in which P-AttEnc is trained in [6, 7, 8]-way and tested in a 2-way, where the 2-way are unseen drivers. The accuracy of P-AttEnc is significantly lower than that of the previous experiment because P-AttEnc attempts to classify unseen drivers that do not appear in the training data. However, we can observe that the accuracy increases with the increase of ’shot’, and it also increases with the increase of drivers in the training episodes, which indicates that P-AttEnc needs more categories during training to better classify unseen drivers.

\begin{figure}[htbp]
    \centering
    \includegraphics[width=0.5\linewidth]{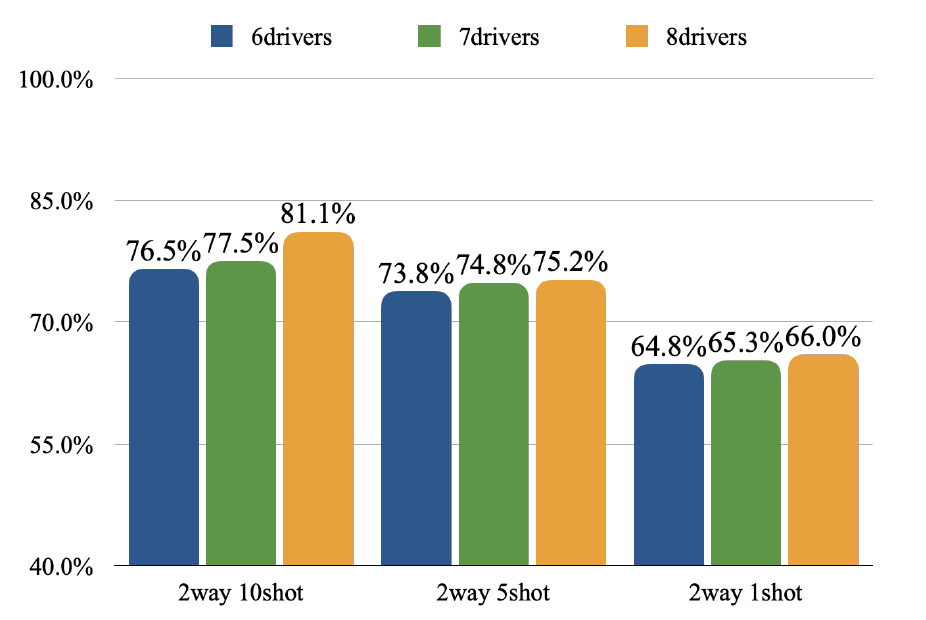}
    \caption{Results of experiment II in stage 2. }
    \label{fig:placeholder}
\end{figure}

\section{Conclusion and Future Works}

In this paper, a highly accurate and efficient driver identification model is proposed, the attention-based encoder (\textit{AttEnc}), which classifies fixed-length time series of vehicle dynamics to identify drivers. It has the highest accuracy compared to the ARNet, CoLSTM\_Att, and CoGRU\_Att models and performs 44\% to 79\% faster than the compared models because it significantly reduces, on average, 87.6\% of the model parameters. For the data shortage issue, the proposed \textit{P-AttEnc }can identify drivers using limited training data. We observed that more "shots" allows \textit{P-AttEnc }to achieve a higher accuracy, but more "ways" slightly decreases the accuracy of \textit{P-AttEnc}.

Driver identification techniques can be applied to many applications, which will increase their demand in the near future. For example, it can be used to prevent theft of vehicles, home vehicles and car sharing, fleet driver management, and UBI vehicle insurance. Moreover, we can use a few-shot learning method to address the issue of data shortages, which can help driver identification technology develop under the restrictions of low cost and data shortages. We believe that the proposed method can be applied in the following areas:

\begin{itemize}
\item  Anti-theft System

\item  User-Based Insurance

\item  Fleet Management

\item  Personalized Service
\end{itemize}
This study has shortcomings, and we will continue to improve it in the future. The datasets that are used are too small to reflect the real world, so larger datasets will be applied in the future. It is hoped that our method can be applied to any data source that is related to vehicle dynamics. During the experiments, we found that GPS data alone cannot be classified to allow driver identification, and it must be combined with an IMU data source (three-dimensional acceleration) to give a good prediction. Finally, the accuracy of \textit{P-AttEnc is }reduced when the "way" is higher, so different few-shot learning methods will be attempted in the future. 
\section*{Acknowledgments}
This work is partially supported by MOST (MOST 109-2410-H-006-117, 110-2410-H-006-064) and the Center for Innovative FinTech Business Models of National Cheng Kung University, sponsored by MOE, Taiwan, R.O.C.

\end{document}